# Development and Validation of a Deep Learning Model for Prediction of Severe Outcomes in Suspected COVID-19 Infection

Varun Buch[1*], Aoxiao Zhong[2,3*], Xiang Li[2*], Marcio Aloisio Bezerra Cavalcanti Rockenbach[1], Dufan Wu[2], Hui Ren[2], Jiahui Guan[4], Andrew Liteplo[5], Sayon Dutta[5], Ittai Dayan[2], Quanzheng Li[1,2]

[1] MGH & BWH Center for Clinical Data Science, Boston, MA

[2] Department of Radiology, Massachusetts General Hospital, Boston, MA

[3] School of Engineering and Applied Sciences, Harvard University, Boston, MA

[4] NVIDIA, Santa Clara, CA, USA

[5] Department of Emergency Medicine, Massachusetts General Hospital

[*] Varun Buch, Aoxiao Zhong and Xiang Li contribute equally to this work.

Corresponding author: **Quanzheng Li**; li.quanzheng@mgh.harvard.edu; (617)-643-9481; 55 Fruit Street White 427 Boston, MA 02114

## Abstract

**Importance:** COVID-19 patient triaging with predictive outcome of the patients upon first present to emergency department (ED) is crucial for improving patient prognosis, as well as better hospital resources management and cross-infection control.

**Objective:** Development and validation of a deep learning-based predictive model for severe COVID-19 outcomes.

**Design:** We trained a deep feature fusion model to predict patient outcomes, where the model inputs were EHR data including demographic information, co-morbidities, vital signs and laboratory measurements, plus patient's CXR images. The model output was patient outcomes defined as the most insensitive oxygen therapy required. For patients without CXR images, we employed Random Forest method for the prediction. Predictive risk scores for COVID-19 severe outcomes ("CO-RISK" score) were derived from model output and evaluated on the testing dataset, as well as compared to human performance.

**Participants:** The study's dataset (the "MGB COVID Cohort") was constructed from all patients presenting to the Mass General Brigham (MGB) healthcare system from March 1st to June 1st, 2020. ED visits with incomplete or erroneous data were excluded. Patients with no test order for COVID or confirmed negative test results were excluded. Patients under the age of 15 were also excluded. Finally, electronic health record (EHR) data from a total of 11060 COVID-19 confirmed or suspected patients were used in this study. Chest X-ray (CXR) images were also collected from each patient if available.

**Main Measure:** The most insensitive oxygen therapy required in 24/72 hours for the patients admitted to ED. The therapies include room air (RA), low-flow oxygen (LFO), high-flow oxygen/non-invasive ventilation (HFO/NIV), or mechanical ventilation (MV). Relationship between CO-RISK score and patient's management decision (discharge, admit to standard care or admit to intensive care) and mortality were also analyzed in this study.

**Results:** The CO-RISK score achieved area under the Curve (AUC) of predicting MV/death (i.e. severe outcomes) in 24 hours of 0.95, and 0.92 in 72 hours on the testing dataset. The model shows superior performance to the commonly used risk scores in ED (CURB-65 and MEWS). Comparing with physician's decisions, CO-RISK score has demonstrated superior performance to human in making ICU/floor decisions.

**Conclusions and Relevance:** The developed prediction model has demonstrated accurate and robust performance across different sites and during different disease prevalence, indicating its potential usefulness in clinical practice, especially for estimating the baseline risk of patients during initial presentation in the ED, or where the patients cannot get their COVID-19 diagnosis immediately.

1. **Introduction**

First identified in the Hubei province of China in December 2019, the severe acute respiratory syndrome coronavirus 2 (SARS-CoV-2)[1] has spread globally in a matter of months to trigger a pandemic of unprecedented scale and severity. By November 2020, a combination of high infectivity and pathogenicity has resulted in more than 1.3 million deaths resulting from at least 53.7 million known infections[2]. In human hosts, the virus results in Coronavirus Disease 2019 (COVID-19), which is characterized by a flu-like illness in mild cases, multi-lobar pneumonia, acute respiratory distress syndrome (ARDS) and multi-organ failure in the most severe of cases[3].

As the pandemic is going to prevail for months, it will continuously challenge the health system on its allocation of resource (e.g. mechanical ventilator, oxygen, ICU beds and experts in respiratory intensive care unit). Currently, most hospitals can triage COVID-19 like patients to clinic or emergency department according to their initial presentation. However, it is less effective on identifying those with worse prognosis (needing advanced oxygen therapy, mechanical ventilator, or have a higher risk of death) based on patient's first present to the ED. Determining the patient's risk of severe outcome is crucial for two main reasons. Firstly, during times when case numbers are surging, such as outbreaks, resources such as mechanical ventilators, personal protective equipment (PPE) and intensive care unit (ICU) beds are likely to be in short supply[4]. Secondly, admitting a patient with COVID-19 disease into a care facility increases the chances of the condition spreading to vulnerable patients that are already admitted in the hospital, which facilitating the spread of the condition.

A clinical decision support system for helping stratify the severity of COVID-19 in the Emergency Department (ED) would therefore represent significant value to the disease management as well as to the hospital operations, and there have been increasing number of applications leveraging artificial intelligence especially deep learning systems for patient screening, triaging and management[5,6]. Although there are established scores for risk stratification that can be used in ED, such as the CURB-65[7] for acute onset of pneumonia, or the Modified Early Warning System (MEWS) score[8] for more generally acute unwell patient, COVID-19 represents an entirely new human disease with its own characteristics and prognostic signature. Consequently, there have been many early studies attempting to predict COVID-19 progression based on the data captured during the pandemic[9,10]. However, these models have generally used small sample sizes[11], have been limited to a single site for model training and testing[12], and have used patient and data selection methodology that could lead to bias[13]. Furthermore, existing work has not compared model performance against the performance of physicians making similar decisions to the models; thus, it remains unclear whether such models would help or hinder physicians in making COVID-19 management decisions.

In this study, we conduct a large-scale analysis of consecutive patients with suspected COVID-19 infection, presenting to one of five Emergency Departments in the Greater Boston Area of Massachusetts, with the objective of developing and validating a predictive model for severe COVID-19 outcomes. Our approach utilizes the full spectrum of data elements available at initial patient presentation to a physician in the ED, including, patient demographics, vital signs, lab results and imaging studies. This allows us to make a fair comparison between our model and human performance at the task of COVID-19 patient management. Due to the heterogeneity and scale of our input data, our study harnesses artificial intelligence, specifically deep learning, as the principal modelling approach.

## 2. Material and Methods

### 2.1. Data acquisition and cohort selection

The study's dataset (the "MGB COVID Cohort") was constructed from all patients presenting to one of the five Emergency Departments (ED) in the Greater Boston Area of Massachusetts within the Mass General Brigham (MGB) healthcare system from March 1st to June 1st, 2020. All patients were followed-up until June 30th, 2020 and therefore, the minimum observation period was 30 days. ED visits with incomplete or erroneous data, such as missing visit outcome information, contradictory EHR timestamps (e.g. visit ending before visit starting) and unusually long or short visits (typically records created for administrative reasons) were excluded. Patients under the age of 15 were excluded. Patients who were not suspected of having COVID-19, by virtue of not having a SARS-CoV-2 Antigen Polymerase Chain Reaction (PCR) test ordered at the time of the visit or 14 days prior, were excluded. Furthermore, patients with a confirmed negative test any time in the past 14 days were also excluded. The EHR dataset was retrospectively collected from the MGB Enterprise Data Warehouse (EDW). We also retrospectively collected chest X-ray (CXR) images from the clinical Picture Archiving and Communication System (PACS), if the patient had an X-ray scan performed within 24 hours of visit. The study was approved by the institutional review board under data use agreement 2020P000819 and was compliant with the Health Insurance Portability and Accountability Act. Waiver for the need to obtain informed consent was granted. TRIPOD guidelines for reporting of multivariable prediction models were followed[14].

**2.2. Predictors and outcomes**

For each patient included in the cohort, predictor data and outcome data were collected based on the data's temporal relation to the 'ED Decision Point' -- the time point at which an ED physician made a decision about the patient's destination (discharge, admit to standard care or admit to intensive care). Prior to the decision point, predictors were collected including demographic information, co-morbidities, vital signs and laboratory measurements. CXR images both of anterior-posterior (AP) and posterior-anterior (PA) views acquired at the ED visit or up to 24 hours before the visit were also collected. After the decision point, outcomes were collected including the most intensive oxygen therapy required in the next 24/72 hours after decision point: room air (RA), low-flow oxygen (LFO), high-flow oxygen/non-invasive ventilation (HFO/NIV), or mechanical ventilation (MV), patient's management decision (discharge, admit to standard care or admit to intensive care), and mortality. The devices considered in each category of oxygen therapy are listed in supplemental materials. Mortality was captured at 24 hours, 72 hours and 30 days after ED visit.

**2.3. Model development and CO-RISK score transformation**

As the CO-RISK performs patient risk prediction based on learning the non-linear relationship between predictors and outcomes, as a first step we split the whole cohort into training, validation and testing sets. Among five ED sites involved in this study, data collected from two sites were used for training and validation, and the remaining three sites were used for testing the model. EHR data were preprocessed by standard de-identification and missing value imputation via the MissForest algorithm[15]. Patient outcomes were transformed into two (24/72 hours) separate labels with range from 0 to 1. Label values between 0 and 0.75 was related to the most intensive oxygenation therapy the patient had received (0: RA, 0.25: LFO, 0.5: HFO/NIV, 0.75:MV). Label value of 1 indicated that the patient had died. CXR images were preprocessed by automatic series selection to ensure validity and quality of scan, as well as image cropping to 224*224. For the co-analysis of EHR and CXR data from the same patient, CO-RISK employs a deep feature fusion method developed and implemented by us based on the Deep & Cross network architecture[16]. Specifically, binary/categorical features in HER, as well as image features in CXR were firstly transformed into fused dense vectors of real values by embedding and stacking layers, which served as network input. After that, in addition to a classic deep neural network where several fully-connected feed-forward layers are stacked, another cross network is introduced to enforce fusion among features from different sources. The cross network performs explicit feature crossing within its layers by conducting inner products between the original input feature and output from the previous layer, thus increasing the degree of interaction across features. Final output of the model, with the labels described above, is derived from the concatenation of both classic and cross networks. Training of the network took roughly 20 minutes of computational time. For patient with only EHR data available, where no cross-modality fusion is needed, CO-RISK utilizes Random Forest method for the outcome prediction[17].

As both the deep learning and Random Forest methods performed the same task of risk prediction and had the same type of model output (continuous value from 0 to 1, in 24/72 hours), results from them were combined for a single model evaluation: for patients with CXR available, prediction from the deep learning method was used; and for patients with only EHR data, prediction from Random Forest was used. In order to establish the final CO-RISK scores, a cube root transform was applied to the combined results, followed by multiplication by a factor of 100. The cube root transform reduced the skewness of the score distribution[18], while the multiplication made score more readable.

Feature importance of the CO-RISK model was evaluated by permutation importance[19].

where we randomly permutated the value of each feature (i.e. item in EHR data) and recorded the corresponding changes in prediction error. Higher increase in prediction error due to the permutation indicates higher importance of that feature.

**2.4. Clinical score calculation**

Two widely-used clinical scores were selected in this work in order to compare their performance with the new proposed method: CURB-65[7] and MEWS[8]. CURB-65 is a six-points score based on confusion, urea, respiratory rate, blood pressure, and age that can be used to stratify community acquired pneumonia patients into different management groups and perform 30 day mortality risk estimation[7]. MEWS is also a point based system that evaluates vital signs (systolic blood pressure, heart rate, respiratory rate and temperature) and mental state to identify patients at risk for deterioration (ICU admission, cardiorespiratory emergency and death)[8]. There have been no studies validating the use of these scores in COVID-19 patients. Both scores could be calculated using EHR data of each patient. We compare the performance of the proposed CO-RISK model with these two scores by the corresponding Receiver Operating Characteristic (ROC) curves for predicting the necessity for MV in 72 hours, as such prediction is highly needed in ED patient and resource management. Additional comparisons are provided in supplemental materials based on the ROC for 30-days mortality (using CURB-65) and for 30-days composite outcome (using MEWS), similar to the outcomes described in their original literatures.

**2.5. Human performance calculation**

In order to evaluate physician's performance in making clinical decisions, we investigated the scenario where physicians were making ICU (a high dependency care unit) or floor (a normal dependency care unit) assignment at the patient's initial present, which is a major decision to be made at ED. We established the ground truth for ICU/floor assignment based on whether the patient received MV in 72 hours after the ED visit, for the same reason of ED necessity. Then we calculated the physician's performance (sensitivity/specificity) based on the correspondence between patient's ICU/floor decision and 72-hours MV. We also obtained the ROC curve of the CO-RISK score performing the same task based on the ground truth. Finally, we referred to the physicians' performance to calculate the score threshold, by identifying the point on CO-RISK ROC curve that is closest (measured by Euclidean distance) to the physicians' sensitivity and specificity.

3. Result

**Demographics of the MGB COVID Cohort**

A total of 11060 COVID-19 confirmed or suspected patients with EHR data available were included in the cohort (mean age, 57 years [standard deviation 20 years]; 49.4% male; 62.6% White and 13.2% African American). Flow of patients through this study can be found in Figure 1.

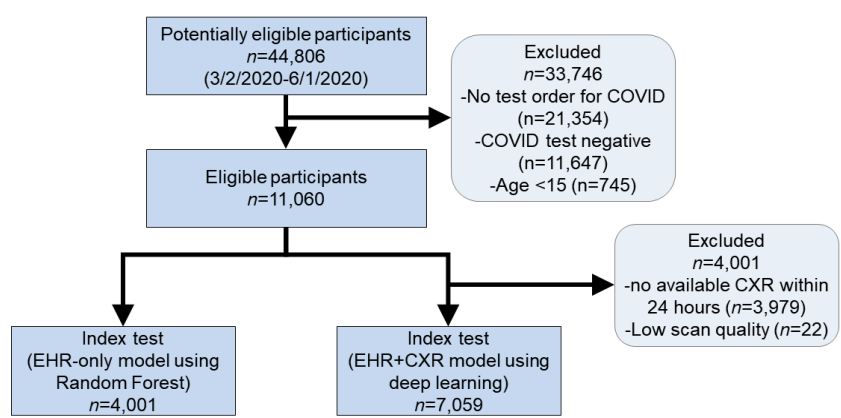

Figure 1: Flow of participants through the study, showing the inclusion/exclusion criteria and the number of subjects available to the index tests.

By the time of ED visit, 30.9% of the included patients were confirmed with positive COVID-19 PCR test. The most common existing comorbidities included hypertension (4426, 40%), diabetes (2029, 18.3%), and obesity (1703, 15.4%), in which morbid obesity accounted for 4.2%. At the initial presentation in ED, 902 patients (8.2%) were febrile (>= 38°C), 1185 (17.0%) had respiratory rate greater than 24 breaths/minute, and 2109 (19.1%) received oxygen therapy at the ED. In the initial laboratory measures, 2782 (31.4%) patients had lymphopenia (<1000*109/L), 239 (4.2%) and 1650 (28.7%) had abnormal liver function (ALT/AST), and 1020 (11.9%) had abnormal kidney function. A full list of patient's existing comorbidities, initial vital signs and initial labs can be found in the corresponding sections in Table 1. Within 24 hours of the ED Decision Point, 3104 (28.1%) of the patient would receive supplemental oxygen (LFO,

HFO/NIV or MV). Within 72 hours, 3681 (33.3%) would receive supplemental oxygen.

Table 1: Characteristics of the MGB cohort, grouped by data sites

| Characteristic | Total (n=11060) | Site1 (n=4556) | Site2 (n=2401) | Site3 (n=998) | Site4 (n=1873) | Site5 (n=1232) |
|---|---|---|---|---|---|---|
| | | Training (n=6957) | | Testing (n=4103) | | |
| **Baseline and demographic** | | | | | | |
| Age, mean (SD), yr. | 56.7 (19.8) | 55.9 (19.2) | 55.9 (20.5) | 59.3 (19.9) | 55.9 (18.8) | 60.0 (21.6) |
| Male, no. (%) | 5463 (49.4) | 2416 (53.0) | 1119 (46.6) | 490 (49.1) | 846 (45.2) | 592 (48.1) |
| Race, no. (%) | | | | | | |
|     Asian | 364 (3.3) | 160 (3.5) | 63 (2.6) | 15 (1.5) | 70 (3.7) | 56 (4.5) |
|     Black or African American | 1463 (13.2) | 441 (9.7) | 252 (10.5) | 263 (26.4) | 437 (23.3) | 70 (5.7) |
|     Hispanic or Latino | 293 (2.6) | 84 (1.8) | 14 (0.6) | 65 (6.5) | 122 (6.5) | 8 (0.6) |
|     Other | 1450 (13.1) | 769 (16.9) | 196 (8.2) | 133 (13.3) | 197 (10.5) | 155 (12.5) |
|     Unavailable | 568 (5.1) | 287 (6.3) | 79 (3.3) | 68 (6.8) | 93 (5.0) | 41 (3.3) |
|     White or Caucasian | 6922 (62.6) | 2815 (61.8) | 1797 (74.8) | 454 (45.5) | 954 (50.9) | 902 (73.2) |
| Smoking, no. (%) | 1165 (10.5) | 468 (10.3) | 370 (15.4) | 99 (9.9) | 155 (8.3) | 73 (5.9) |
| **COVID-19 Positive** | 3417 (30.9) | 1364 (29.9) | 776 (32.3) | 308 (30.9) | 479 (25.6) | 490 (39.8) |
| **Comorbidities, no. (%)** | | | | | | |
| Anemia | 1907 (17.2) | 801 (17.6) | 391 (16.3) | 162 (16.2) | 393 (21.0) | 160 (13.0) |
| Cancer | 1837 (16.6) | 767 (16.8) | 298 (12.4) | 105 (10.5) | 471 (25.1) | 196 (15.9) |
| Cardiovascular disease | 8252 (74.6) | 3492 (76.6) | 1896 (79.0) | 677 (67.8) | 1306 (69.7) | 881 (71.5) |
| Cerebrovascular disease | 655 (5.9) | 295 (6.5) | 119 (5.0) | 51 (5.1) | 133 (7.1) | 57 (4.6) |
| Chronic kidney disease | 1084 (9.8) | 457 (10.0) | 291 (12.1) | 89 (8.9) | 145 (7.7) | 102 (8.3) |
| Respiratory disease | 2397 (21.7) | 980 (21.5) | 636 (26.5) | 223 (22.3) | 366 (19.5) | 192 (15.6) |
| Coagulopathy | 436 (3.9) | 193 (4.2) | 75 (3.1) | 34 (3.4) | 98 (5.2) | 36 (2.9) |
| History of transplant | 201 (1.8) | 114 (2.5) | 9 (0.4) | 6 (0.6) | 66 (3.5) | 6 (0.5) |
| Liver disease | 597 (5.4) | 305 (6.7) | 153 (6.4) | 39 (3.9) | 72 (3.8) | 28 (2.3) |
| Metabolic Disease | 3732 (33.7) | 1573 (34.5) | 936 (39.0) | 336 (33.7) | 578 (30.9) | 309 (25.1) |
| Neurodegenerative disease | 472 (4.3) | 156 (3.4) | 108 (4.5) | 48 (4.8) | 70 (3.7) | 90 (7.3) |
| Pregnancy | 116 (1.0) | 27 (0.6) | 44 (1.8) | 4 (0.4) | 30 (1.6) | 11 (0.9) |
| **Initial vital signs** | | | | | | |
| Temperature, mean (SD), °C | 36.9 (0.7) | 36.7 (0.5) | 36.9 (1.0) | 37.1 (0.5) | 37.1 (0.9) | 37.1 (0.6) |
| SpO2, median (IQR), % | 97 (3.0) | 97 (3.0) | 98 (3.0) | 98 (3.0) | 98 (3.0) | 97 (4.0) |
| Received supplemental oxygen at initial presentation, no. (%) | 2109 (19.1) | 1056 (23.2) | 364 (15.2) | 171 (17.1) | 287 (15.3) | 231 (18.8) |
|     Low flow oxygen | 1584 (14.3) | 769 (16.9) | 290 (12.1) | 128 (12.8) | 210 (11.2) | 187 (15.2) |
|     High flow oxygen or noninvasive ventilation | 292 (2.6) | 141 (3.1) | 49 (2.0) | 30 (3.0) | 33 (1.8) | 39 (3.2) |
|     Mechanical Ventilation | 233 (2.1) | 146 (3.2) | 25 (1.0) | 13 (1.3) | 44 (2.3) | 5 (0.4) |
| Respiratory rate, mean (SD), breaths/min | 20.4 (5.9) | 21.1 (6.1) | 19.6 (6.7) | 19.8 (5.0) | 19.2 (4.5) | 21.3 (5.4) |
| Heart rate, mean (SD), beats/min | 91.4 (19.8) | 92.5 (20.5) | 91.5 (19.9) | 90.7 (18.4) | 90.7 (19.3) | 88.7 (18.7) |
| Systolic BP, mean (SD), mmHg | 138.4 (25.9) | 138.5 (26.2) | 140.2 (26.6) | 136.7 (23.9) | 136.6 (25.7) | 138.3 (24.6) |
| Diastolic BP, mean (SD), mmHg | 78.4 (14.8) | 77.7 (14.7) | 81.2 (15.6) | 79.7 (15.5) | 76.2 (14.0) | 77.8 (13.7) |
| **Initial laboratory measures, median (IQR)** | | | | | | |
| Alanine aminotransferase, U/L | 22 (14-36) | 23 (15-39) | 21 (14-36) | 21 (14-38) | 21 (13-33) | 20 (13-33) |
| Aspartate aminotransferase, U/L | 27 (20-43) | 29 (21-47) | 26 (19-40) | 27 (20-47) | 25 (19-37) | 25 (18-38) |
| C reactive protein, mg/dL | 31.8 (7.7-89.5) | 33.0 (8.1-88.2) | 29.0 (7.3-96.9) | 40.5 (5.8-97.4) | 32.2 (6.5-90.2) | 29.5 (8.5-80.8) |
| Creatinine, mg/dL | 0.9 (0.8-1.2) | 0.9 (0.8-1.2) | 0.9 (0.8-1.2) | 0.9 (0.7-1.2) | 0.9 (0.8-1.2) | 0.9 (0.7-1.2) |
| Ferritin, µg/L | 276 (118-638) | 274 (114-618) | 267 (113-679) | 255 (112-699) | 304 (134-819) | 289 (124-591) |
| GFR, ml/min/1.73m$^2$ | 78 (53-98) | 78 (53-98) | 79 (56-98) | 80 (53-98) | 76 (53-97) | 78 (52-97) |
| Glucose, mg/dL | 116 (100-146) | 116 (100-147) | 119 (102-148) | 119 (104-151) | 112 (96-140) | 115 (102-140) |
| Hemoglobin, g/L | 13.1 (11.6-14.4) | 13.1 (11.5-14.4) | 13.2 (11.9-14.4) | 13.1 (11.8-14.4) | 12.7 (11.0-14.0) | 13.4 (12.0-14.6) |
| Lactate, mmol/L | 1.5 (1.1-2.3) | 1.5 (1.1-2.3) | 1.5 (1.1-2.4) | 1.6 (1.1-2.3) | 1.5 (1.1-2.3) | 1.5 (1.2-2.2) |
| Lactate dehydrogenase, U/L | 255 (200-351) | 260 (203-351) | 242 (198-324) | 266 (206-386) | 265 (202-369) | 242 (191-340) |
| Lymphocyte, *10$^9$/L | 1.3 (0.9-2.0) | 1.3 (0.8-1.9) | 1.4 (1.0-2.0) | 1.4 (0.9-2.0) | 1.4 (0.8-2.0) | 1.3 (0.9-1.9) |
| Neutrophils, *10$^9$/L | 5.3 (3.7-7.9) | 5.5 (3.8-8.1) | 5.3 (3.8-7.7) | 5.3 (3.7-7.6) | 5.2 (3.5-7.7) | 5.3 (3.7-7.9) |
| Platelet, *10$^9$/L | 225 (175-283) | 224 (173-284) | 225 (177-280) | 232 (182-289) | 231 (179-292) | 216 (172-275) |
| Potassium, mmol/L | 4.0 (3.7-4.4) | 4.1 (3.8-4.4) | 4.0 (3.7-4.3) | 4.0 (3.7-4.4) | 4.0 (3.7-4.4) | 4.1 (3.8-4.4) |
| Sodium, mmol/L | 139 (136-141) | 138 (135-140) | 140 (137-142) | 138 (135-140) | 139 (136-141) | 139 (136-141) |
| WBC, *10$^9$/L | 7.8 (5.8-10.5) | 7.9 (5.9-10.7) | 7.9 (5.9-10.5) | 7.8 (5.9-10.2) | 7.7 (5.7-10.2) | 7.7 (5.7-10.5) |

| Table 1. Characteristics of the MGB cohort, grouped by data sites | | | | | | |
|---|---|---|---|---|---|---|
| **Outcome, no (%)** | | | | | | |
| Oxygen therapy in 24h (highest O$_2$ required in period) | | | | | | |
|    Low flow | 2355 (21.3) | 1085 (23.8) | 429 (17.9) | 176 (17.6) | 390 (20.8) | 275 (22.3) |
|    High flow or noninvasive ventilation | 244 (2.2) | 83 (1.8) | 64 (2.7) | 27 (2.7) | 28 (1.5) | 42 (3.4) |
|    Mechanical ventilation | 506 (4.6) | 307 (6.7) | 55 (2.3) | 32 (3.2) | 80 (4.3) | 32 (2.6) |
| Oxygen therapy in 72h (highest O$_2$ required in period) | | | | | | |
|    Low flow | 2662 (24.1) | 1250 (27.4) | 454 (18.9) | 202 (20.2) | 444 (23.7) | 312 (25.3) |
|    High flow or noninvasive ventilation | 370 (3.3) | 128 (2.8) | 85 (3.5) | 38 (3.8) | 53 (2.8) | 66 (5.4) |
|    Mechanical ventilation | 650 (5.9) | 371 (8.1) | 93 (3.9) | 41 (4.1) | 105 (5.6) | 40 (3.2) |
| Admitted to ICU | 821 (7.4) | 408 (9.0) | 155 (6.5) | 56 (5.6) | 165 (8.8) | 37 (3.0) |
| Admitted to Floor | 5693 (51.5) | 2470 (54.2) | 1042 (43.4) | 509 (51.0) | 977 (52.2) | 695 (56.4) |
| Death | 719 (6.5) | 268 (5.9) | 143 (6.0) | 69 (6.9) | 143 (7.6) | 96 (7.8) |
| **Outcome, follow-up in 7 days who discharged from ED, no (%)** | | | | | | |
| Represent to ED | 321 (7.2) | 124 (7.5) | 93 (7.9) | 22 (5.1) | 41 (5.7) | 41 (8.3) |
| Readmission | 293 (6.5) | 117 (7.0) | 77 (6.6) | 23 (5.4) | 39 (5.4) | 37 (7.5) |

Based on the training/testing data scheme as introduced in section 2.3, there were 6957 and 4103 patients assigned to the training and testing dataset, respectively. There was no significant difference in age and gender between training and testing dataset. More intubation and mechanical ventilation were observed in the training samples.

Among all patients, 7056 patients also have chest X-ray images. Thus, EHR+CXR data from 7056 patients were analyzed by the deep learning (Deep & Cross network) method, the rest of 4004 patients with only EHR data available were analyzed by Random Forest. Outputs from both methods were then combined and transformed to 24/72 hours risk scores for every patient in the cohort.

**Prediction performance of CO-RISK and important features**

ROC curves of the CO-RISK model for predicting whether patients in the testing dataset will need MV or die within 24/72 hours, which is a major goal of CO-RISK model, are shown in Figure 2. We also obtained the ROC curves for predicting the need for other oxygen therapies, which are provided in supplemental materials. Area under the Curve (AUC) of predicting MV/death in 24h was 0.95 (95%CI, 0.92-0.96). For MV in 72h, the AUC was 0.92 (95%CI, 0.90-0.94). Based on feature permutation importance analysis, it was found that the following patient characteristics were of the topmost importance for the prediction: two vital signs (SPO2 and respiratory rate), the oxygen device that the patient is using upon ED visit, as well as patient's age. Secondly important are the following laboratories: lactate, lactate dehydrogenase, C reactive protein, and neutrophils. The last group that is considered as important for the predictions includes systolic/diastolic blood pressure, as well as other laboratories (aspartate aminotransferase, glomerular filtration rate, platelet, troponin T, glucose, d-dimer, creatinine and white blood cell count).

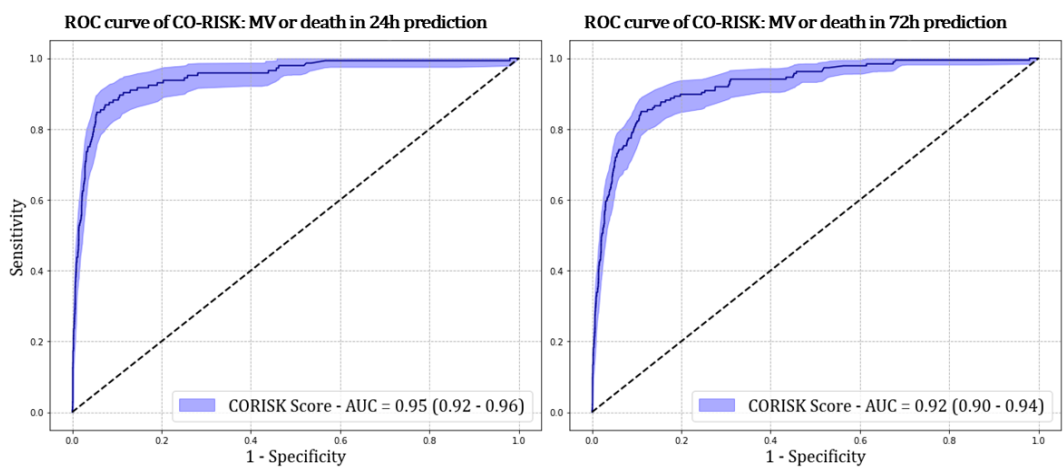

Figure 2: ROCs of predicting the need for MV or death within 24 and 72 hours. Error bars show the 95% CI calculated from 1000 bootstrapping.

**Comparison with other clinical scores**

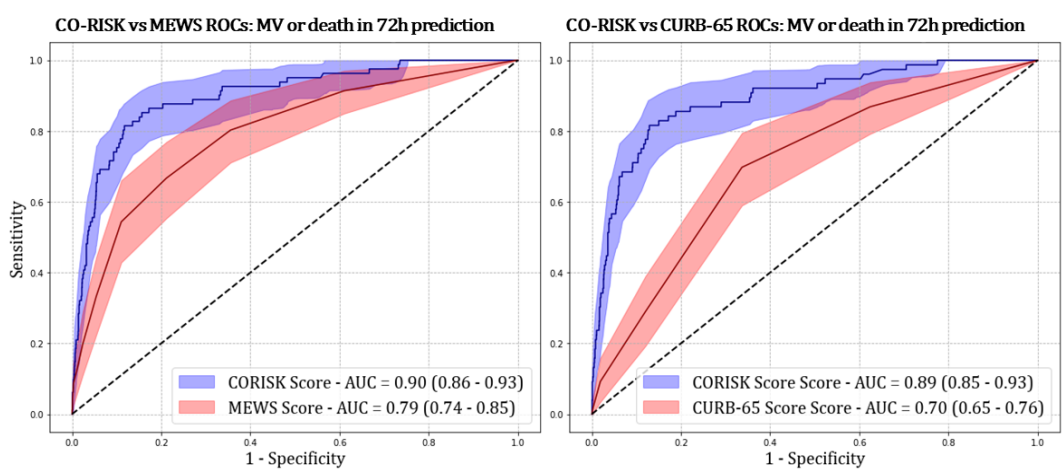

Figure 3: Performance comparison in predicting the necessity for MV or death in 72 hours. Left: CO-RISK vs MEWS. Right: CO-RISK vs CURB-65.

Out of the 4103 patients in the testing dataset, 2048 (49.9%) of them had all the necessary information to calculate the MEWS score, 1748 (42.6%) had information needed for the CURB-65 score. Thus, we compared the ROC curves for making the 72-hours MV or death prediction by CO-RISK model and the two clinical scores on the corresponding subsets of the testing dataset, as plotted in Figure 3. Because of the difference in data used, ROC curves of CO-RISK are different in both plots as well as in Figure 2. Comparing with MEWS score which had an AUC of 0.79 (0.74-0.85 95%CI), CO-RISK reached AUC of 0.90 (0.86-0.93 95%CI). Comparing with CURB-65 which had an AUC of 0.70 (0.65-0.76 95%CI), CO-RISK reached AUC of 0.89 (0.85-0.93 95%CI). Comparisons of 30-days morality (CO-RISK vs. CURB-65) and composite outcome (CO-RISK vs. MEWS) showed similar superior performance from CO-RISK, which are provided in the supplemental materials.

**Comparison with physician's performance**

Based on the scheme for evaluating physician's performance as introduced in section 2.5, we obtained the sensitivity and specificity in deciding ICU/floor assignment by physicians and CO-RISK model in the testing dataset. ROC curve of CO-RISK score for this task and the identified threshold (red asterisk and the corresponding error bar), along with physician's performance (green dot and error bar), are shown in Figure 4. For further comparison, we also moved the threshold of CO-RISK to the same level of sensitivity as the physicians (0.672), where CO-RISK achieved specificity (95% CI) of 0.953 (0.946 - 0.960) and physician's specificity was 0.966 (0.960 - 0.971).

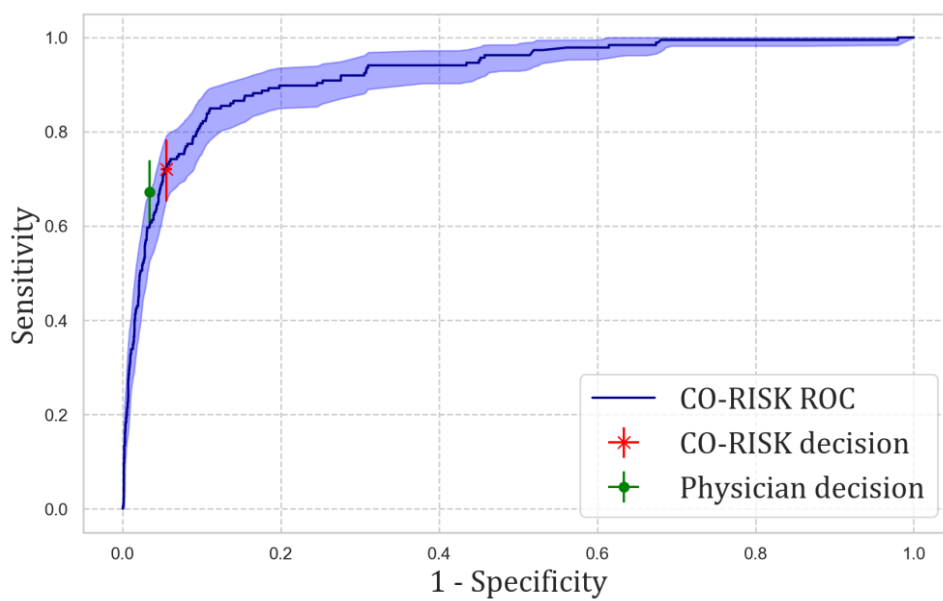

Figure 4: CO-RISK compared to physicians in deciding patient destination after ED visit: ICU vs floor. CO-RISK decision point was identified using the threshold discussed in section 2.5.

As we have also collected the 30-days morality data in the MGB COVID cohort, we analyzed patients' survival stratification based on the physician's decision for ICU/floor or discharged using Kaplan-Meier survival curve, as shown in Figure 5(a). We also obtain the K-M curve using CO-RISK score, where patients were stratified as 'Low', 'Medium' and 'High' risk of death based on the threshold derived from training set, as in Figure 5(b). The thresholds were established in order to best match physician's decisions in the training set: patient discharged, admitted to floor, or admitted to the ICU. Based on the K-M curves of CO-RISK score, it was found that the mortality of patients in the high-risk group was significantly higher than in the medium- and low-risk groups (38.24% vs 6.52% and 0.45%; p<0.001).

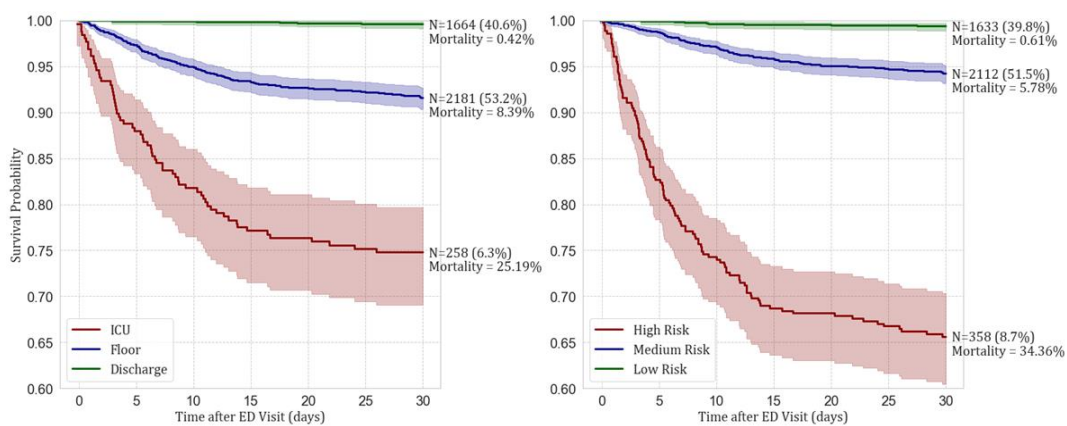

Figure 5: Left: 30-day K-M curve for patients based on physician's decision. Right: K-M curve based on CO-RISK score.

## 4. Discussion

**Clinical implications of CO-RISK score**

The CO-RISK score was developed to predict short-term (24/72h) oxygen requirements, rather than just a binary outcome (e.g. death). Therefore, the score is potentially more clinically useful, since it is more actionable and directly points to the physician what kind of treatment and resources the patient might require. In the study of comparison with physician's decisions, CO-RISK score has demonstrated superior performance to human in making ICU/floor decisions. As shown in Figure 5, patients determined by the model as "high risk" have a much worse survivability, comparing with patients sent to ICUs in reality. Thus, in a scenario where resources are limited, the score can help physicians stratify patients and better plan the use of available equipment and hospital beds. The model could also be deployed for management of patients in areas unfamiliar with the clinical condition. The threshold suggested by the model provides a reference decision point which could be adjusted to accommodate available resources.

**Adaptiveness of CO-RISK score**

In the current experiment setting, we split the MGB COVID cohort into training and testing dataset based on different sites and found that CO-RISK can adapt well to the changes in hospitals. To further investigate the adaptiveness of CO-RISK to different time period, which also reflect prevalence of COVID, we split the data based on different time periods within the study timeframe. The CO-RISK model was trained by data from all hospitals during March to April 2020. It was then tested on three test sets: test set I included ED visits between May 1st and May 10th, test set 2 between May 11th and May 20$^{th}$, test set 3 between May 21st and May 31st. As the number of new cases fluctuated during the pandemic, the rate of positive case varied accordingly. In the MGB COVID cohort, the portion of COVID-19 positive patients presented to ED was 41.1% during 3/3 - 4/30, but dropped to 33.1%, 18.5% and 13.1% during 5/1 - 5/10, 5/11 - 5/20, and 5/21 - 5/31, respectively. Performance of CO-RISK score for the prediction of oxygen therapies in 24/72 hours using the new data split show similar AUCs with the results we reported above, indicating that CO-RISK can also adapt well to different COVID prevalence. In addition, this train/test data split is equivalent to a prospective data collection, indicating the feasibility of CO-RISK to be applied prospectively. Detailed analysis of the prediction performance can be found in supplemental materials.

**Applicability of CO-RISK score**

CO-RISK model was developed without relying on COVID-19 diagnosis status (e.g. from PCR test), which made it suitable for application in ED where the patients cannot get their COVID-19 diagnosis immediately upon visit. One of the most critical challenges in COVID-19 research is the development and testing of therapeutics aiming to treat infected patients that are experiencing a severe illness. The CO-RISK score could be used at initial presentation in the ED to calculate baseline risk of future deterioration. Thus, the CORISK score could be used to assess the therapeutic benefit of candidate therapies in clinical trials. Furthermore, the score would allow more targeted enrollment of patients likely to experience a more severe illness, affording a better risk/reward trade-off for patients receiving experimental therapy.

**Limitations of CO-RISK score**

We compared the performance of CO-RISK score with physicians on dispositioning patients to ICU/floor, where the result showed similar performance. Currently, we are unable to account for unmeasured confounders in the decision-making process. However, it is less likely and unethical to conduct randomized controlled clinical trials, especially in the stressing environment of ED at a stretching moment in the pandemic. During preliminary attempts of deploying CO-RISK in clinical workflow, we found certain patients are missing key clinical variables and CXRs, which prohibited the model from running. Such missing data pattern is informative and could be related to the patient's clinical condition when presenting to ED. For instance, we found that relatively severe patients were more likely to have CXR scans, and physicians would order a blood panel or ferritin based on their clinical judgement.

The MGB COVID cohort was established based on patients from five teaching hospitals within the same healthcare system in Boston, which share similar technical infrastructure in electronic health record (EHR), data storage and ED protocols. The generalization and scale up of CO-RISK score should be cautious and tailored to the hospital-specific context, depending on the existing information infrastructure, data availability and healthcare provider's adoption.

## 5. Conclusion

In this study of using EHR and CXR data for predicting of severe outcomes of COVID-19, we developed a deep learning-based risk score which demonstrated high sensitivity and specificity for predicting patient outcome and making clinical decisions. Further research is necessary for a prospective deployment in ED and integration into clinical workflow. We are also investigating the feasibility of using federated learning scheme to incorporate additional sites into CO-RISK score development, in order to establish an international framework of COVID-19 patient risk stratification.

# Development and Validation of a Deep Learning Model for Prediction of Severe Outcomes in Suspected COVID-19 Infection

# Supplemental Materials

**Validation of the CO-RISK on different disease prevalence periods: additional scheme for training/testing data splitting**

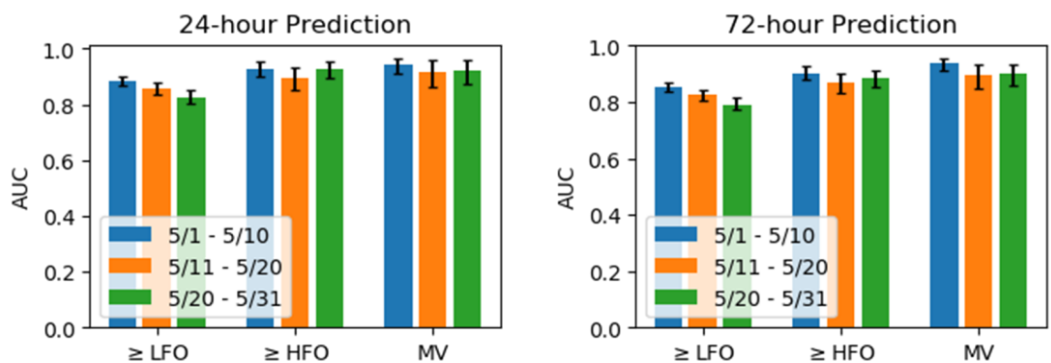

Supplemental Figure 1: AUCs of oxygen therapy prediction in 24/72 hours using CO-RISK, on the three testing sets covering three different time periods. AUCs were calculated for the need of oxygen therapy more aggressive than LFO, HFO/NIV, or MV. The error bars show the 95% confidence interval calculated from 1,000 bootstrapping experiments.

As shown in Supplemental Figure 1, similar performance can be achieved for predicting all the three oxygen therapy types in 24/72 hours using CO-RISK. Model performance are slightly lower in more recent testing dataset (green bars, May 21st to May 31st, 2020), as the patient population in this testing set was more deviated from the training dataset (March to April 2020).

**List of oxygen devices involved in each oxygen therapy types**

Room air (RA): no oxygen device needed

Low flow oxygen (LFO) devices: Nasal cannula, Simple mask, Oxymask, Oxygen conserving device, Blow-by, Pulse dose device, Aerosol mask

High flow oxygen devices(HFO) / Non-invasive ventilation(NIV): High flow nasal cannula, Face tent, High flow face mask, Bag-valve Mask, Non-rebreather mask, T-Piece, Venturi mask, Partial rebreather mask, Bi-PAP, CPAP, Transtracheal catheter

Mechanical ventilation (MV): Ventilator

**Additional patient outcome prediction by CO-RISK and comparison with other clinical scores**

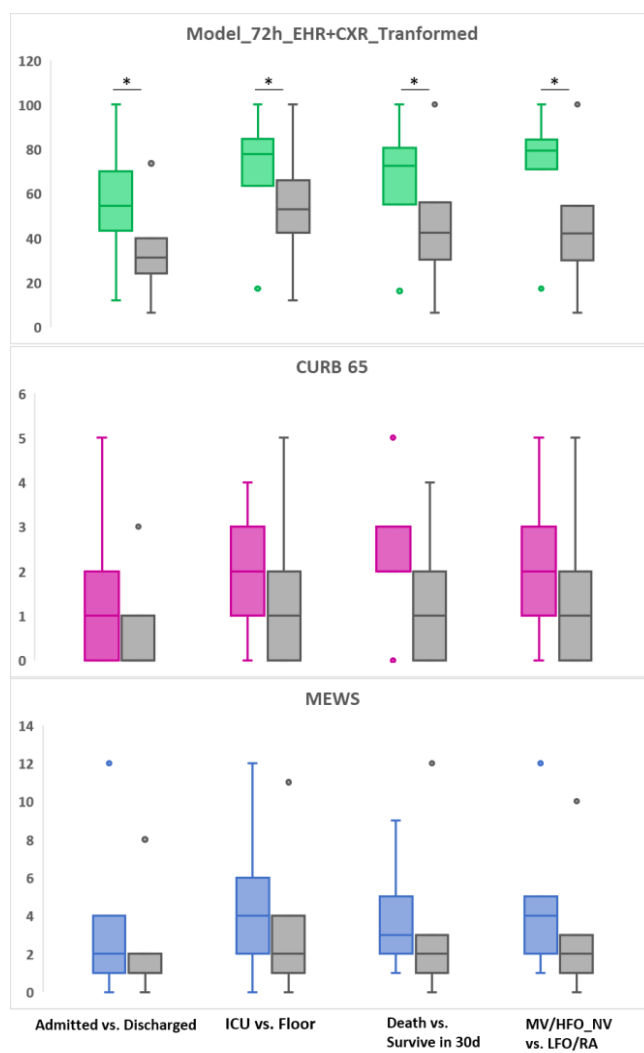

Supplemental Figure 2: Boxplots for CO-RISK (top panel), CURB-65 (middle panel), and MEWS (bottom panel) scores in differentiating four different types of patient outcomes: "Admitted vs. Discharged", "ICU vs. Floor", "Death vs. Survive in 30 days", and "MV/HFO vs. LFO/RA". In each boxplot, minimum sample value, 25th percentile, 50th percentile, 75th percentile, and maximum sample value are provided.

In the univariate analysis as shown in Supplemental Figure 2, the CO-RISK score can significantly differentiate admitted vs. discharged patients (54[43-70] vs.31[24-40],

p<0.001), patients admitted to ICU versus floor (78[63-85] vs. 53[42-66], p<0.001), patients deceased versus survived in 30 days (73[55-81] vs. 42[30-56], p<0.001), as well as patients on mechanical ventilation or high flow oxygen versus low flow oxygen or room air (79[71-84] vs. 42[30-55], p<0.001). Although CURB-65 and MEWS also show statistically difference, these two scores have more overlap between groups.

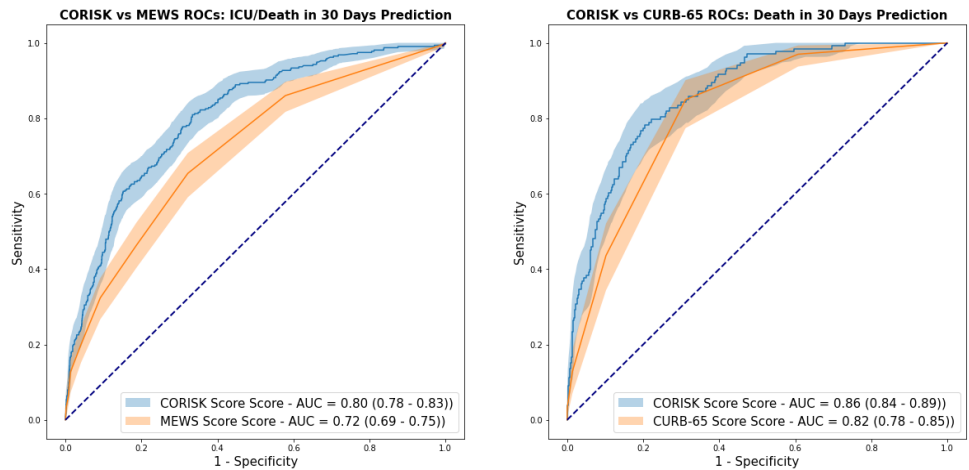

Supplemental Figure 3: ROC curves of ICU/Death prediction using CO-RISK and MEWS score (left), and 30-days death prediction using CO-RISK and CURB-65 scores (right).

As MEWS score was used to prompt earlier intervention on the ward and earlier transfer to ICU[1,2], we specifically compared the performance between CO-RISK and MEWS for predicting ICU/Death. As shown by the ROC curves in Supplemental Figure 3, CO-RISK achieved an AUC of 0.80 (0.78 - 0.83 95% CI), where MEWS achieved an AUC of 0.72 (0.69 - 0.75 95% CI). Similarly, as CURB-65 score was designed and validated for predicting mortality in pneumonia and lung infection[3], we specifically compared the performance between CO-RISK and CURB-65 scores in predicting 30-day morality. CO-RISK achieved an AUC of 0.86 (0.84 - 0.89 95% CI), where CURB-65 achieved an AUC of 0.82 (0.78 - 0.85 95% CI).

It is worth mentioning that those results were obtained using the CO-RISK score trained for 72-hours outcome prediction, which means that the model was not trained for these specific outcomes (admission/discharge, ICU/floor, or 30-day morality). Even so, CO-RISK achieved comparable or even better performance than the other clinical scores in predicting the patient outcomes they original designed to predict.